\def\year{2020}\relax
\documentclass[letterpaper]{article} % DO NOT CHANGE THIS
\usepackage{aaai20}  % DO NOT CHANGE THIS
\usepackage{times}  % DO NOT CHANGE THIS
\usepackage{helvet} % DO NOT CHANGE THIS
\usepackage{courier}  % DO NOT CHANGE THIS
\usepackage[hyphens]{url}  % DO NOT CHANGE THIS
\usepackage{graphicx} % DO NOT CHANGE THIS
\usepackage{graphicx}  % DO NOT CHANGE THIS
\usepackage{amsmath}
\usepackage{amssymb}
\usepackage{bbold}
\DeclareMathOperator*{\argmax}{arg\,max}
\usepackage{xcolor}
\nocopyright
\usepackage{booktabs}       % professional-quality tables

\title{Frustratingly Easy \textit{Natural} Question Answering}
\author{Lin Pan\thanks{Equal Contribution.}, Rishav Chakravarti\footnotemark[1], Anthony Ferritto, Michael Glass,\\
 \Large{\textbf{Alfio Gliozzo, Salim Roukos, Radu Florian, Avirup Sil\thanks{ Corresponding author.}}}\\ % All authors must be in the same font size and format. Use \Large and \textbf to achieve this result when breaking a line
IBM Research AI\\
Yorktown Heights, NY\\
\{panl, rchakravarti, mrglass, gliozzo, roukos, raduf, avi\}@us.ibm.com \\
aferritto@ibm.com % Anthony: my email is @ibm.com unlike most other people's
}

\definecolor{darkgreen}{rgb}{0.0, 0.2, 0.13}

\newcommand{\bertqa}{{\sc BERT-for-QA}}

\newcommand{\bertbase}{BERT$_{\text{BASE}}$}
\newcommand{\bertlarge}{BERT$_{\text{LARGE}}$}
\newcommand{\best}[1]{\textbf{#1}}
\newcommand{\citet}[1]{\citeauthor{#1}~\shortcite{#1}}

\begin{document}

\maketitle

\begin{abstract}
Existing literature on Question Answering (QA) mostly focuses on algorithmic novelty, data augmentation, or increasingly large pre-trained language models like XLNet and RoBERTa. Additionally, a lot of systems on the QA leaderboards do not have associated research documentation in order to successfully replicate their experiments. In this paper, we outline these algorithmic components such as Attention-over-Attention, coupled with data augmentation and ensembling strategies that have shown to yield state-of-the-art results on benchmark datasets like SQuAD, even achieving super-human performance. Contrary to these prior results, when we evaluate on the recently proposed Natural Questions benchmark dataset, we find that an incredibly simple approach of transfer learning from BERT outperforms the previous state-of-the-art system trained on $4$ million more examples than ours by $1.9$ F1 points. Adding ensembling strategies further improves that number by $2.3$ F1 points. 

\end{abstract}

\section{Introduction}
\label{sec:intro}

A relatively new field in the open domain question answering (QA) community is machine reading comprehension (MRC) which aims to read and comprehend a given text, and then answer questions based on it. MRC is one of the key steps for natural language understanding (NLU). MRC also has wide applications in the domain of conversational agents and customer service support. Among the most widely worked on MRC benchmark datasets are the Stanford SQuAD v1.1 \cite{Rajpurkar_2016} and v2.0 \cite{rajpurkar2018know} datasets. Recent MRC research has explored transfer learning from large pre-trained language models like BERT \cite{Devlin2018BERTPO} and XLNet \cite{xlnet} which have solved the tasks in less than a year since their inception. Hence, we argue that harder benchmark MRC challenges are needed. In addition, the SQuAD datasets both suffer from observational bias: the datasets contain questions and answers provided by annotators who have read the given passage first and then created a question given the context. Other datasets like NarrativeQA \cite{kovcisky2018narrativeqa} and HotpotQA \cite{yang2018hotpotqa} are similarly flawed.

In this paper, we focus on a new benchmark MRC dataset called Natural Questions (NQ) \cite{Kwiatkowski2019NaturalQA} which does not possess the above bias. The NQ queries were sampled from Google search engine logs according to a variety of handcrafted rules to filter for ``natural questions'' that are potentially answerable by a Wikipedia article.  This is a key differentiator from past datasets where observation bias is a concern due to the questions having been generated \textit{after} seeing an article or passage containing the answer \cite{Kwiatkowski2019NaturalQA}. Also, systems need to extract a short and a long answer (paragraphs which would contain the short answer). The dataset shows a human upper bound of 76\% on the short answer and 87\% on the long answer selection tasks. Since the task has been recently introduced and is bias-free, the authors claim that matching human performance on this task will require significant progress in natural language understanding.

% Our algorithmic novelties include building on top of a SOTA BERT model: Attention-over-attention (AoA) \cite{cui_2017} + BERT. This adds additional attention layers on top of a matured BERT transformer which already performs self-attention \cite{Vaswani_2017}. This adds additional attention layers on top of a matured BERT transformer which already performs self-attention \cite{Vaswani_2017} . We further improve the BERT layers to perform a linear combination similar to ELMo embeddings \cite{Peters_2018}. \pan{maybe just: 
The contributions of our paper include:
\begin{itemize}
    \item \textbf{Algorithmic novelties:} We add an Attention-over-attention (AoA) \cite{cui_2017} layer on top of BERT during model finetuning, which gives us the best single model performance on NQ. We also perform a linear combination of BERT output layers instead of using the last layer only. Additionally, we show empirically that an incredibly simple transfer learning strategy of finetuning the pre-trained BERT model on SQuAD first and then on NQ can nearly match the performance of further adding the complex AoA layer.
\item \textbf{Smarter Data Augmentation:} We show that a simple but effective data augmentation strategy that shuffles the training data helps outperform the previous state-of-the-art (SOTA) system trained on $4$ million additional synthetically generated QA data.
\item \textbf{Ensembling Strategies:} We describe several methods that can combine the output of single MRC systems to further improve performance on a leaderboard. Most previous work that obtains ``super-human"\footnote{\citet{rajpurkar2018know} note that human performance is likely somewhat underestimated.} performance on the leaderboard fail to outline their ensembling techniques.
\end{itemize}

\section{Related Work}
\label{sec:relatedwork}
Most recent MRC systems are predominantly BERT-based as is evident on leaderboards for SQuAD v1.1 and v2.0, HotpotQA and Natural Questions. ``Super-human'' results are achieved by adding additional components on top of BERT or BERT-like models such as XLNet. Among them, \textit{XLNet + SG-Net Verifier} \cite{zhang2019sg} adds a syntax layer, and \textit{BERT + DAE + AoA} adds an AoA component as shown on the SQuAD leaderboard.

Another common technique is data augmentation by artificially generating more questions to enhance the training data. \citet{albert-synth-data}, an improvement over \citet{alberti2019bert}, combine models of question generation with answer extraction and filter results to ensure round-trip consistency. This technique helped them gather an additional 4 million synthetic training examples which provides SOTA performance on the NQ task. 

Top submissions on the aforementioned leaderboards are usually \textit{ensemble} results of single systems, yet the underlying ensemble technique is rarely documented. Even the most popular system, \textit{BERT + N-Gram Masking + Synthetic Self-Training (ensemble)} \cite{Devlin2018BERTPO}, does not provide their ensemble strategies. In this paper, we describe our recipe for various ensemble strategies together with algorithmic improvements and data augmentation to produce SOTA results on the NQ dataset.

% parikh, bert-baseline, alberti-acl : qa models

% aoa: models
\section{Model Architecture}
\label{sec:model-arch}

% \rishav{Perhaps we should call this section "Model Architecture" rather than model improvements and start out by describing the basis BertForQA in a tiny bit more detail (citing the original paper) maybe borrowing ideas from section 3.2 of the EMNLP CFO paper?}
% \pan{Sounds good, so maybe 3 subsections for this section: BertForQA, AoA, Linear combo?}
% \rishav{Yup: 3 sections sounds good}
In this section, we first describe \bertqa, the model our system is built upon, and two algorithmic improvements on top of it. (1) Attention-over-Attention (AoA) \cite{cui_2017}, as an attention mechanism, combines \emph{query-to-document} and \emph{document-to-query} attentions by computing a document-level attention that is weighted by the importance of query words. This technique gives SOTA performance on SQuAD. (2) Inspired by the success of ELMo \cite{Peters_2018}, we use a linear combination of all the BERT encoded layers instead of only the last layer.

\subsection{BERT-for-QA} 
Given a token sequence $\mathbf{X} = [x_1, x_2,\ldots,x_T]$, BERT, a deep Transformer \cite{Vaswani_2017} network, outputs a sequence of contextualized token representations $\mathbf{H}^L = [\mathbf{h}_1^L, \mathbf{h}_2^L,\ldots,\mathbf{h}_T^L]$.
\begin{equation*}
\mathbf{h}_1^L,\ldots,\mathbf{h}_T^L = BERT(x_1,\ldots,x_T)    
\end{equation*}
% In our work, we report results with \bertlarge{}, which consists of $24$ Transformer layers ($L=24$), each with $16$ heads and $\mathbf{h}_t^L\in \mathbb{R}^{1024}$ while \bertbase{} ($L=12$, each layer with $12$ heads and $\mathbf{h}_t^L\in \mathbb{R}^{768}$) is used for intermediate parameter tuning experiments.
\bertlarge{} consists of $24$ Transformer layers ($L=24$), each with $16$ heads and $\mathbf{h}_t^L\in \mathbb{R}^{1024}$ while \bertbase{} is smaller, ($L=12$, each layer with $12$ heads and $\mathbf{h}_t^L\in \mathbb{R}^{768}$). 
As an important preprocessing step for BERT, special markup tokens  {\tt[CLS]} and {\tt[SEP]} are added; one to the beginning of the input sequence and the other to the end. In cases like MRC, where there are two separate input sequences, one for the question and the other for the given context, an additional {\tt[SEP]} is added in between the two to form a single sequence.

{\sc BERT-for-QA} adds three dense layers followed by a \emph{softmax} on top of BERT for answer extraction: $\boldsymbol{\ell}_b = softmax(\mathbf{W}_1 \mathbf{H}^L)$, $\boldsymbol{\ell}_e = softmax(\mathbf{W}_2 \mathbf{H}^L)$ and $\boldsymbol{\ell}_a = softmax(\mathbf{W}_3 \mathbf{h}_{[CLS]}^L)$, where $\mathbf{W}_1$, $\mathbf{W}_2 \in \mathbb{R}^{1\times 1024}$, $\mathbf{W}_3 \in \mathbb{R}^{5\times 1024}$, $\mathbf{H}^L \in \mathbb{R}^{N\times 1024}$, and $\mathbf{h}_{[CLS]}^L \in \mathbb{R}^{1024}$. $\boldsymbol{\ell}_b^t$ and $\boldsymbol{\ell}_e^t$ denote the probability of the $t^{th}$ token in the sequence being the answer beginning and end, respectively. These three layers are trained during the finetuning stage. The NQ task requires not only a prediction for short answer beginning/end offsets, but also a (containing) longer span of text that provides the necessary context for that short answer. Following prior work from \citet{alberti2019bert}, we only optimize for short answer spans and then identify the bounds of the containing HTML span as the long answer prediction\footnote{The candidate long answer HTML spans are provided as part of the preprocessed data for NQ.}. We use the hidden state of the {\tt[CLS]} token to classify the answer type $\in[short\_answer, long\_answer, yes, no, null\_answer]$, so $\boldsymbol{\ell}_a^y$ denotes the probability of the $y^{th}$ answer type being correct. Our loss function is the averaged cross entropy on the two answer pointers and the answer type classifier:
%\begin{equation*}
%\mathcal{L} = - \frac{\sum\limits_{t=1}^T \mathbb{1} (\mathbf{b}_t) \log %\boldsymbol{\ell}_{b}^{t}+\sum\limits_{t=1}^T \mathbb{1} (\mathbf{e}_t) %\log \boldsymbol{\ell}_{e}^{t}+\sum\limits_{y=1}^Y \mathbb{1} %(\mathbf{a}_y) \log \boldsymbol{\ell}_{a}^{y}}{3},
%\end{equation*}
\begin{align*}
    \mathcal{L}_{NQ} = - \frac{1}{3}\left(\sum_{t=1}^T \mathbb{1}(\mathbf{b}_t) \log \boldsymbol{\ell}_{b}^{t} \right. &+ \sum_{t=1}^T \mathbb{1} (\mathbf{e}_t) \log \boldsymbol{\ell}_{e}^{t} \\
&+ \left. \sum_{y=1}^Y \mathbb{1} (\mathbf{a}_y) \log \boldsymbol{\ell}_{a}^{y}\right),
\end{align*}

% \begin{equation}
% \mathcal{L}_{NQ} = - \frac{1}{3}(\sum_{t=1}^T \mathbb{1}(\mathbf{b}_t) \log \boldsymbol{\ell}_{b}^{t} + \sum_{t=1}^T \mathbb{1} (\mathbf{e}_t) \log \boldsymbol{\ell}_{e}^{t} \\
% +\sum_{y=1}^Y \mathbb{1} (\mathbf{a}_y) \log \boldsymbol{\ell}_{a}^{y})
% \end{equation}

% \begin{eqnarray*}
% \mathcal{L}_{NQ} & = & - \frac{1}{3}\left(\sum_{t=1}^T \mathbb{1} (\mathbf{b}_t) \log \boldsymbol{\ell}_{b}^{t}\right. \\
% & & \qquad+ \sum_{t=1}^T \mathbb{1} (\mathbf{e}_t) \log \boldsymbol{\ell}_{e}^{t} \\
% & & \qquad+ \left.\sum_{y=1}^Y \mathbb{1} (\mathbf{a}_y) \log \boldsymbol{\ell}_{a}^{y}\right),
% \end{eqnarray*}

where $\mathbb{1}(\mathbf{b})$ and $\mathbb{1}(\mathbf{e}$) are one-hot vectors for the ground-truth beginning and end positions, and $\mathbb{1}(\mathbf{a}$) for the ground-truth answer type.
During decoding, the span over \emph{argmax} of $\boldsymbol{\ell}_b$ and \emph{argmax} of $\boldsymbol{\ell}_e$ is picked as the predicted short answer. 
% \avi{Pan are you okay with the equation? I formatted it.} \pan{yeah, looks good. I changed the beginning and ending parentheses to bigger ones.}

\subsection{Attention-over-Attention}
AoA was originally designed for cloze-style question answering, where a phrase in a short passage of text is removed in forming a question. Let $\mathbf{Q}$ be a sequence of question tokens $[\mathbf{q}_1,\ldots,\mathbf{q}_m]$, and $\mathbf{C}$ a sequence of context tokens $[\mathbf{c}_1,\ldots,\mathbf{c}_n]$. AoA first computes a attention matrix:
\begin{equation}\label{eq1}
\mathbf{M} = \mathbf{C} {\mathbf{Q}^T},     
\end{equation}
where $\mathbf{C}\in \mathbb{R}^{n\times h}$, $\mathbf{Q}\in \mathbb{R}^{m\times h}$, and $\mathbf{M}\in \mathbb{R}^{n\times m}$. In our case, the hidden dimension is $h = 1024$. Next, it separately performs on $\mathbf{M}$ a column-wise \emph{softmax} $\mathbf{\alpha} = softmax(\mathbf{M}^T)$ and a row-wise \emph{softmax} $\mathbf{\beta} = softmax(\mathbf{M})$. Each row $i$ of matrix $\mathbf{\alpha}$ represents the document-level attention regarding $\mathbf{q}_i$ (\emph{query-to-document} attention), and each row $j$ of matrix $\mathbf{\beta}$ represents the query-level attention regarding $\mathbf{c}_j$ (\emph{document-to-query} attention). To combine the two attentions, $\mathbf{\beta}$ is first row-wise averaged:
\begin{equation}\label{eq2}
\mathbf{\beta} = \frac{1}{n}\sum_{j=1}^n {\beta_j}     
\end{equation}
The resulting vector can be viewed as the average importance of each $\mathbf{q}_i$ with respect to $\mathbf{C}$, and is used to weigh the document-level attention $\mathbf{\alpha}$.
\begin{equation}\label{eq3}
\mathbf{s} = \mathbf{\alpha}^T {\mathbf{\beta}^T}     
\end{equation}
The final attention vector $\mathbf{s}\in \mathbb{R}^N$ represents document-level attention weighted by the importance of query words.

In our work, we use AoA by adding an \emph{two-headed} AoA layer into the BERT-for-QA model and this layer is trained together with the answer extraction layer during the finetuning stage. Concretely, the combined question and context hidden representation $\mathbf{H}^L$ from BERT is first separated to $\mathbf{H}^Q$ and $\mathbf{H}^C$ \footnote{Superscript $L$ is dropped here for notation convenience; we use the last layer $L=24$ from the BERT output.}, followed by \emph{two} linear projections of $\mathbf{H}^Q$ and $\mathbf{H}^C$ respectively to $\mathbf{H}_i^Q$ and $\mathbf{H}_i^C$, $i \in \{1, 2\}$:
\begin{equation}\label{eq4}
\mathbf{H}_i^Q = \mathbf{H}^Q {\mathbf{W}_i^Q},    
\end{equation}
\begin{equation}\label{eq5}
\mathbf{H}_i^C = \mathbf{H}^C {\mathbf{W}_i^C},     
\end{equation}
where $\mathbf{H}^Q$, $\mathbf{H}_i^Q \in \mathbb{R}^{M\times 1024}$; $\mathbf{H}^C$, $\mathbf{H}_i^C \in \mathbb{R}^{N\times 1024}$; and $\mathbf{W}_i^Q, \mathbf{W}_i^C \in \mathbb{R}^{1024\times 1024}$. Therefore, the AoA layer adds about 2.1 million parameters on top of BERT which already has 340 million. Next, we feed $\mathbf{H}_1^C$ and $\mathbf{H}_1^Q$ into AoA calculation specified in Equation \eqref{eq1} to \eqref{eq3} to get the attention vector $\mathbf{s}_1$ for head $1$. The same procedure is applied to $\mathbf{H}_2^Q$ and $\mathbf{H}_2^C$ to get $\mathbf{s}_2$ for head $2$. Lastly, $\mathbf{s}_1$ and $\mathbf{s}_2$ are combined with $\boldsymbol{\ell}_b$ and $\boldsymbol{\ell}_e$ respectively via two weighted sum operations for answer extraction.

% \anthony{Does AoAconcat need to be references above ?} \pan{yes, I'll describe both aoa2h and aoaconcat} \pan{EDIT: actually given the aoaconcat result, I'll probably only describe aoa2h, for which I'm running expts using 2 epochs, and the median result comparison will be between bert-for-qa and aoa2h}
\subsection{BERT Layer Combination}
So far, we have described using the last layer from the BERT output $[\mathbf{h}_1^L,\ldots,\mathbf{h}_n^L]$ as input to downstream layers. We also experiment with combining all the BERT output layers into one representation. Following \citet{Peters_2018}, we create a trainable vector $\mathbf{v}\in \mathbb{R}^L$ and apply \emph{softmax} over it, yielding $\mathbf{w} = softmax(\mathbf{v})$. The output layers are linearly combined as follows:
\begin{equation*}
\mathbf{h}_i^{\prime} = \sum_{l=1}^L w_l\mathbf{h}_i^l  
\end{equation*}
$\mathbf{v}$ is jointly trained with parameters in BERT-for-QA. $\mathbf{h}_i^{\prime}$ is then used as input to the final answer extraction layer.

\section{Model Training}
\label{sec:training}

Our models follow the now common approach of starting with the pre-trained BERT language model and then finetune over the NQ dataset with an additional QA sequence prediction layer as described in previous section.  As mentioned in \cite{alberti2019bert}, we also find it helpful to run additional task specific pre-training of the underlying BERT language model before starting with the finetuning step with the target NQ dataset.  The following two subsections discuss different pre-training and data augmentation strategies employed to try and improve the overall performance of the models. Note that unless we specify otherwise, we are referring to the ``large'' version of BERT.

\subsection{Pre-Training}
\label{subsec:pretraining}

We explore three types of BERT parameter pre-trainings prior to finetuning on the NQ corpus:\\
% \textbf{1. BERT with Whole Word Masking (WWM)} is one of the default BERT pre-trained models
    % made available by Google research\footnote{\url{https://github.com/google-research/bert}} 
    % that follows a nearly identical strategy to the original BERT model \footnote{\url{https://github.com/google-research/bert}} pre-training (henceforth, PT) \pan{remove henceforth?} proposed in \cite{Devlin2018BERTPO}, but uses masks across whole word phrases (rather than word pieces). \pan{
    \begin{enumerate}
    \item \textbf{BERT with Whole Word Masking (WWM)} is one of the default BERT pre-trained models
    % made available by Google research\footnote{\url{https://github.com/google-research/bert}} 
    that has the same model structure as the original BERT model, but masks whole words instead of word pieces for the Masked Language Model pre-training task.
    \item \textbf{BERT with Span Selection Pre-Training (SSPT)} uses an unsupervised auxiliary QA specific task proposed by \citet{Anon_2019} to further train the BERT model. 
    % made available by Google research.  
    The task generates synthetic cloze style queries by masking out terms (named entities or noun phrases) in a sentence.  Then answer bearing passages are extracted from the Wikipedia corpus using BM25 based information retrieval \cite{Robertson_theprobabilistic}.  This allows us to pre-train all layers of the BERT model including the answer extraction weights by training the model to extract the answer term from the selected passage.
   \item \textbf{BERT-for-QA with SQuAD 2.0} finetunes BERT on the supervised task of SQuAD 2.0 as initial pre-training. The intuition is that this allows the model to become more domain and task aware than vanilla BERT.
    %  \avi{or basic or transformer layers (not to confuse with Bert base)} 
    % network as well as the finetuning layers. The intuition is that a model that is able to do QA well on one dataset would probably be better than using just a large pre-trained language model like vanilla BERT.  \pan{how about: \textbf{BERT-for-QA with SQuAD 2.0} first finetunes BERT on the supervised task of SQuAD 2.0 as initial pre-training. The intuition is that a model which is already domain and task aware would probably be a better starting point than vanilla BERT (and then remove the next sentence}. In this case we use SQuAD 2.0 \cite{rajpurkar2018know}. 
    \citet{alberti2019bert} similarly leverage SQuAD 1.1 to pre-train the network for NQ. However, we found better results using SQuAD 2.0, likely because of SQuAD 2.0's incorporation of unanswerable questions which also exist in NQ.
    \end{enumerate}

In our future work, we intend to explore the effect of these pre-trainings on additional language models including RoBERTa \cite{roberta} and XLNet. 

\subsection{Data Augmentation}
\label{subsec:data-aug}

As noted in a number of works such as \cite{DBLP:journals/corr/abs-1809-10735},  and \cite{DBLP:journals/corr/abs-1804-00720}, model performance in the MRC literature has benefited from finetuning the model with labeled examples from either human annotated or synthetic data augmentation from similar tasks (often with the final set of mini batch updates relying exclusively on data from the target domain as described in the transfer learning tutorial by \citet{ruder-etal-2019-transfer}). In fact,  \citet{albert-synth-data} achieve prior SOTA results for the NQ benchmark by adding 4 million synthetically generated QA examples. In this paper, we similarly try to introduce both synthetically generated as well as human labelled data from other related MRC tasks during NQ training.

\subsubsection{Synthetic Data: Sentence Order Shuffling (SOS)}
The SOS strategy shuffles the ordering of sentences in the paragraphs containing short answer annotations from the NQ training set.  The strategy was attempted based on the observation that preliminary Bert-for-QA models showed a bias towards identifying candidate short answer spans from earlier in the paragraph rather than later in the paragraph (which may be a feature of how Wikipedia articles are written and the types of answerable questions that appear in the NQ dataset).  This is similar in spirit to the types of perturbations introduced by \citet{Zhan2019EnsembleBW} for SQuAD 2.0 based on observed biases in the SQuAD dataset. Note that this strategy is much simpler than the genuine text generation strategy employed by \citet{albert-synth-data} to produce the previous SOTA results for NQ which we intend to explore further in future work. 

\subsubsection{Data from other MRC Tasks}
We attempt to leverage human annotated data from three different machine reading comprehension (MRC) datasets for data augmentation:
\begin{enumerate}
    \item \textbf{SQuAD 2.0} - \raisebox{-.5ex}{\textasciitilde}130,000 crowd sourced question and answer training pairs derived from Wikipedia paragraphs.
    \item \textbf{NewsQA} \cite{newsqa} - \raisebox{-.5ex}{\textasciitilde}100,000 crowd sourced question and answer training pairs derived from news articles.
    \item \textbf{TriviaQA} \cite{DBLP:journals/corr/JoshiCWZ17} - \raisebox{-.5ex}{\textasciitilde}78,000 question and answers authored by trivia enthusiasts which were subsequently associated with wikipedia passages (potentially) containing the answer.
\end{enumerate}

\subsubsection{Augmentation Data Sampling}

Our simple BERT-for-QA model takes about 20 hours to train a single epoch on the roughly 300,000 NQ training examples using a system with 2 Nvidia\textsuperscript{\textregistered} Tesla\textsuperscript{\textregistered} P100 GPUs. Introducing augmentation data, therefore, can (1) increase training time dramatically and (2) begin to overshadow the examples from the target NQ dataset. So we try two sampling strategies for choosing human annotated MRC examples from past datasets: (1) random and (2) based on question-answer similarity to the NQ dataset.  

% Introducing all three human annotated MRC datasets during training would (1) increase training time dramatically and (2) begin to overshadow the ~300,000 question + article pairs from the target NQ dataset.  In fact, it already takes 20 hours to train a single epoch on just the NQ training data using a system with 2 Nvidia\textsuperscript{\textregistered} Tesla\textsuperscript{\textregistered} P100 GPUs.  We, therefore, tried sub-sampling the augmentation data using two different strategies: (1) random and (2) based on question-answer similarity to the NQ dataset.  

For similarity based sampling, we follow a strategy similar to \citet{DBLP:journals/corr/abs-1809-06963}.  Specifically, we train a BERT-for-Sequence-Classification model using the Huggingface PyTorch implementation of BERT \footnote{\url{https://github.com/huggingface/pytorch-transformers.}}. The model accepts question tokens (discarding question marks since those do not appear in NQ) as the first text segment and short answer tokens (padded or truncated to 50 to limit maximum sequence length) as the second text segment. The model is trained with cross entropy loss to predict the source dataset for the question-answer pair using the development set from the three augmentation candidate datasets as well as target NQ development set. 

Once trained, the predicted likelihood of an example being from the NQ dataset is calculated for all question-answer pairs from the three augmentation candidate training datasets and used to order the examples by similarity for the purposes of sampling\footnote{The BERT-for-Sequence-Classification model achieves ~90\% accuracy at detecting the dataset source for a given query-answer pair.}. As would be expected, the most ``similar'' question-answer pairs were from SQuAD 2.0 (\raisebox{-.5ex}{\textasciitilde}80\% of the sampled data came from SQuAD 2.0) since the task is well aligned with the NQ task while TriviaQA question-answer pairs tended to be least ``similar'' (only \raisebox{-.5ex}{\textasciitilde}9.5\% of the sampled data came from TriviaQA).   
\section{Experiments}
\label{sec:experiments}

\subsection{Dataset}
\label{subsec:dataset}
The NQ dataset provides 307,373 queries for training, 7,830 queries for development, and 7,842 queries for testing (with the test set only being accessible through a public leaderboard submission).  
% Queries were sampled from Google search engine logs according to a variety of handcrafted rules to filter for ``natural questions'' that are potentially answerable by a Wikipedia article.  This is a key differentiator from past datasets where observation bias is a concern on account of the questions having been generated \textit{after} seeing article/passage containing the answer \cite{Kwiatkowski2019NaturalQA}. 

For each question, crowd sourced annotators also provide start and end offsets for short answer spans\footnote{Instead of short answer spans, annotators have marked ~1\% of the questions with a simple Yes/No.  We leave it as future work to detect and generate answers for these types of queries.} within the Wikipedia article, if available, as well as long answer spans (which is generally the most immediate HTML paragraph, list, or table span containing the short answer span), if available \cite{Kwiatkowski2019NaturalQA}.

Similar to other MRC datasets such as SQuAD 2.0, the NQ dataset forces models to make an attempt at ``knowing what they don't know'' by requiring a confidence score with each prediction. The evaluation script\footnote{The evaluation script is provided by Google at \url{https://github.com/google-research-datasets/natural-questions}.}, then calculates the optimal threshold at which the system will ``choose" to provide an answer.  The resulting F1 scores for Short Answer (SA) and Long Answer (LA) predictions are used as our headline metric.

The ``partial un-answerability'' and ``natural generation'' aspects of this dataset along with the recency of the task's publication make it an attractive dataset for evaluating model architecture and training choices (with lots of headroom between human performance and the best performing automated system).

The training itself is carried out using the Huggingface PyTorch implementation of BERT which supports starting from either \bertbase{} or \bertlarge{}. 
% \subsection{Competitors}
% \avi{mention the SOTA: Alberti-ACL2019 paper, the bert-baseline paper, others: Parikh etc.}

% \rishav{This section may need to be re-visited to avoid repetition based on what gets written in section \ref{sec:model-arch} about the model.  Also, hopefully that section clarifies that our final models start from BERT Large and that BERT Base is only used for intermediate parameter tuning experiments.} \anthony{Do we need to mention training on SQUAD before NQ?  I see it's mentioned later, not sure if a quick mention of it here is needed too.} \rishav{I'm rolling that into item `2` below about model pre-training; so we should be OK?}

% We find that the training process is sensitive to things like hyperparameters and pre-training task selection.  As a result, this section describes (1) our hyperparameter optimization strategies, (2) model pre-training,  and (3) data augmentation during the fine-tuning phase. 

\subsection{Hyperparameter Optimization}
\label{subsec:hypopt}
The primary hyperparameter settings for the models discussed in the Model Architecture section are derived from \cite{alberti2019bert} with the exception of the following:

% \begin{table}[]
% \centering
% \begin{tabular}{ll}
% \hline
% Stride Length & $F1$ \\ \hline \hline
% 128 \cite{alberti2019bert} & 53.57 \\ \hline
% 192 & \textbf{54.44} \\ \hline
% 256 & 53.49 \\ \hline
% \end{tabular}
% \caption{The effect of varying  Short Answer F1 on the NQ dev set with varying strides for the sliding window used to split large text sequences.}
% \label{tab:f1-stride}
% \end{table}

\begin{enumerate}
  \item \textbf{Stride} - Following the implementation of the BERT-for-QA model in \cite{Devlin2018BERTPO}, we accommodate BERT's pre-trained input size constraint of 512 tokens by splitting larger sequences into multiple spans over the Wikipedia article text using a sliding window. We experiment with multiple stride lengths to control for both experiment latency (shorter strides results in a larger number of spans per article) as well as F1 performance.
  
  \item \textbf{Negative Instance Sub-Sampling} - Another consequence of splitting each Wikipedia article into multiple spans is that most spans of the article do not contain the correct short answer (only ~65\% of the questions are answerable by a short span and, of these, ~90\% contain a single correct answer span in the article with an average span length of only ~4 words). As a result, there is a severe imbalance in the number of positive to negative (i.e. no answer) spans of text. The authors of \cite{alberti2019bert} address the imbalance during training by sub-sampling negative instances at a rate of 2\%.
  
  We emulate this sub-sampling behavior when generating example spans for answerable questions.  However, based on the observation that our preliminary \bertbase{} models tended to be overconfident for unanswerable questions, we vary the sampling rate between answerable and unanswerable questions. 
  
  \item \textbf{Batch Size \& Learning Rate} - These parameters were tuned for each experiment using the approach outlined in \cite{smith2018disciplined} where we evaluate a number of batch sizes and learning rates on a randomly selected 20\% subset of the NQ training and development data. During experimentation, we did find that slight changes in learning rate can have a couple of points impact on the final F1 scores. Further work is needed to improve robustness of learning rate selection.
  
\end{enumerate}

\subsection{Ensembling}
% Table showcasing experimental results on "train/test" split
In addition to optimizing for single model performance, in this section we outline a number of strategies that we investigated for ensembling models as is common for top ranking leaderboard submissions in MRC\footnote{The top ranking submissions for SQuAD 2.0, TriviaQA, and HotpotQA are all ensemble models as of this paper's writing.}.
In order to formally compare approaches we partition the NQ dev set into ``dev-train'' and ``dev-test'' by taking the first three dev files for the ``train'' set and using the last two for the ``test'' set (the original dev set for NQ is partitioned into 5 files for distribution).  This yields ``train'' and ``test'' sets of 4,653 and 3,177 examples (query-article pairs) respectively.  

For each ensembling strategy considered we search for the best k-model ensemble over the ``train'' set and then evaluate on the ``test'' set.  For these experiments we use $k=4$ as this is the number of models that we can decode in 24 hours on a Nvidia\textsuperscript{\textregistered} Tesla\textsuperscript{\textregistered} P100 GPU, which is the limit for the NQ leaderboard.

% We begin by outlining our core strategy that underlies the approaches we have investigated.  Using this core strategy we investigate a baseline approach of ensembling multiple versions of the same model trained with different seeds in addition to a number of search, normalization, and aggregation strategies and the impact they have on F1 performance.

We examine two types of ensembling experiments: (i) ensembling the same model trained with different seeds and (ii) ensembling different model architectures and \mbox{(pre\textendash)training} data.  Ensembling the same model trained on different seeds attempts to smooth the variance to produce a stronger result.  On the other hand ensembling different models attempts to find models that may not be the strongest individually but harmonize well to produce strong results.

To generate the ensembled predictions for an example, we combine the top-20 candidate long and short answers from each system in the ensemble\footnote{We empirically find that considering 20 is better than considering fewer candidates (e.g. 5 or 10).}.  To combine systems we take the arithmetic mean\footnote{We have experimented with other approaches such as median, geometric mean, and harmonic mean; however these are omitted here as they resulted in much lower scores than arithmetic mean.} of the scores for each long and short span predicted by at least one system.  For spans which are only predicted by a subset of models, a score of zero is imputed for the remaining models. The predicted long/short span is then the span with the greatest arithmetic mean.

\subsubsection{Seed experiments}
We investigate ensembling the best single model, selected as the model with greatest sum of short and long answer F1 scores, trained with $k$ unique seeds.

\subsubsection{Multiple Model Ensembling Experiments}

In our investigation of ensembling multiple models we greedy and exhaustive search strategies for selecting models from a pool of candidate models consisting of various configurations described in the Model Training and Model Architecture sections. The candidate pool also contains multiple instances of the same model training and architecture configuration, but with different learning rates (as mentioned in the previous section, we found that slight changes in learning rate can affect the final performance by a couple of F1 points):

\textbf{Exhaustive Search}
During exhaustive search, we consider all ${n \choose k}$ ensembles of k candidates from our group of n models. After searching all possible ensembles we return two ensembles: (i) the ensemble with the highest long answer F1 score and (ii) the ensemble with the highest short answer F1 score. Given the combinatorial complexity, we limit the search to the top 20 best performing models.  We select the top models using the same approach as in our seed experiments (i.e. the ones with the greatest sum of short and long answer F1 scores).

\textbf{Greedy Search} For the greedy approach we consider all 41 \bertlarge{} models that we had trained during experimentation and greedily build an ensemble of size k from this model set, optimizing for either short or long answer performance.  We refer to the ensembles created in this way as $S$ and $L$ respectively.

We construct $S$ by greedily building $1, 2, ..., k$ model ensembles optimizing for short answer F1.  % using our core strategy.  
In case adding some of the models decreased our short answer performance, we take the first $i \leq k$ models of $S$ which give the highest short answer F1.  The same is done for $L$ when optimizing for long answers.

To build the long answer ensemble (when optimizing for short answer performance), we check to see which subset of $S$ results in the best long answer performance.  More formally we create $L^{\prime} = \argmax_{x \in \mathcal{P}(L)} F1_{L}(x)$ where $F1_{L}(X)$ is the long answer F1 for the ensemble created with the models in $X$.  A corresponding approach is used to create $S^{\prime}$ when optimizing for long answers.

Finally, we join the predictions for short and long answers together by taking the short answer and long answer predictions from our short and long answer model sets respectively.  If for an example a null long answer is predicted, we also predict a null short answer regardless of what $S^{\prime}$ predicted as there are no short answers for examples which do not have a long answer in NQ \cite{Kwiatkowski2019NaturalQA}.

\textbf{Duplicate Answer Span Aggregation} A consequence of splitting large paragraphs into multiple overlapping is that, often, a single system for a single example will generate identical answer spans multiple times in its top 20 predictions. 
% We only keep the maximum scoring span in these case 
In order to produce a unique prediction score for each answer span from each system, we experiment with the following aggregation strategies on the vector $P$ of scores for a given answer span. 
% (1) \textbf{Max} ($\max_{i=1}^{|P|} P_{i}$), (2) \textbf{Reciprocal Rank Sum} ($\sum_{i=1}^{|P|} P_{i} * \frac{1}{i}$), (3) \textbf{Exponential Sum}($\sum_{i=1}^{|P|} P_{i} * \beta^{i - 1}$,  for some constant $\beta$ (we use $\beta = 0.5)$), and (4) \textbf{Noisy-Or} ($1 - \prod_{i=1}^{|P|} (1 - P_{i})$).
\begin{itemize}
    \item \textbf{Max} $ = \max_{i=1}^{|P|} P_{i}$
    \item \textbf{Reciprocal Rank Sum} $= \sum_{i=1}^{|P|} P_{i} * \frac{1}{i}$
    \item \textbf{Exponential Sum} $ = \sum_{i=1}^{|P|} P_{i} * \beta^{i - 1}$ for some constant $\beta$ (we use $\beta = 0.5)$.
    \item \textbf{Noisy-Or} $ = 1 - \prod_{i=1}^{|P|} (1 - P_{i})$
\end{itemize}{}

For the last three strategies\footnote{Using un-normalized versions of sum and noisy-or causes dramatic deterioration.} (reciprocal rank sum, exponential sum, and noisy-or), we additionally experiment with score normalization using a logistic regression model that was trained to predict top 1 precision based on the top score\footnote{Though we experimented with additional input features such as query length and mean score across top 20, we omit results as performance does not improve over simple logistic regression.} using the ``dev-train'' examples.  We use the scikit-learn \cite{scikit-learn} implementation of logistic regression (with stratified 5-fold cross-validation to select the L2 regularization strength).  
\section{Results}
\label{sec:results}

\begin{figure}
\begin{center}
\fbox{\includegraphics[width=\columnwidth]{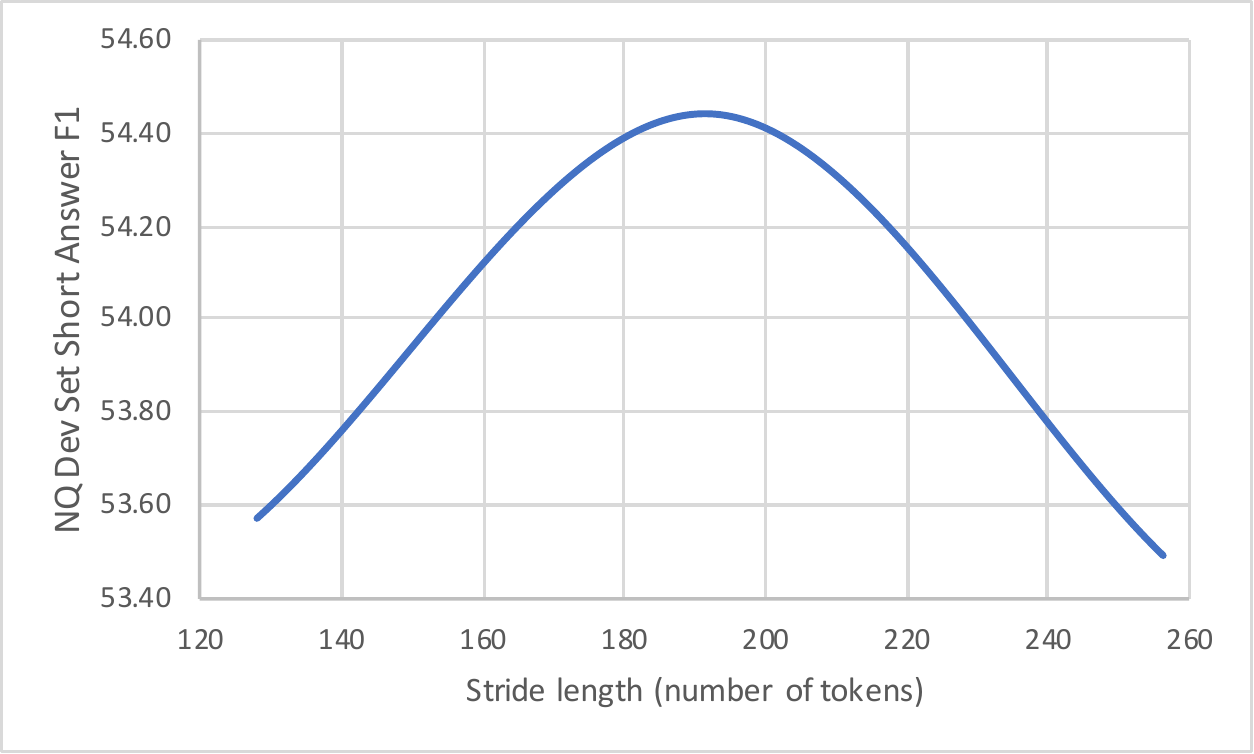}}
\end{center}
\caption{Effect of stride length (in tokens) on the NQ Short Answer Dev Set F1 Performance}
\label{fig:strides}
\end{figure}

\subsubsection{Stride} Rather than using a stride length of 128 tokens as was done by \cite{Devlin2018BERTPO} and \cite{alberti2019bert}, we find that increasing the stride to 192 improves the final F1 score while also reducing the number of spans and, thus, the training time. See figure \ref{fig:strides} for experimental results showing a ~0.9\% gain by increasing the stride length to 192 on some preliminary Bert-for-QA models. 
  
Further increases seem to deteriorate the performance which may be a function of the size of the relevant context in Wikipedia articles, though additional work is required to better explore context size selection approaches given the document text.

\subsubsection{Negative Instance Sub-Sampling} 
As per table \ref{tab:subsampling}, performance initially improves as we sample negative instances at slightly higher rates than the 2\% level used in \cite{alberti2019bert}, but eventually begins to deteriorate when the sampling rate is increased too much. Performance can be improved further by sampling at a slightly \textit{lower} rate of 1\% for answerable questions and at \textit{higher} rate of 4\% for un-answerable questions. Overall, this change provides a boost of ~0.8\% in SA F1 over the setting used in \cite{alberti2019bert} on preliminary \bertbase{}-for-QA models.

\subsubsection{Pre-Training} 
As per table \ref{tab:pretraining-dataaug}, pre-training on SQuAD 2.0 from the WWM  model provides the best single BERT-for-QA model on the target NQ dataset. So we use apply this pre-training strategy to the additional model architectures discussed earlier: AoA and Layer Combo.

\subsubsection{Model Architecture}
Given our best pre-training strategy of the WWM model on SQuAD 2.0, we show in table \ref{tab:pretraining-dataaug} that adding the AoA layer during the finetuning stage of our target dataset of NQ yields the best single model performance. Linearly combining the BERT output layers shows a slight improvement over BERT-for-QA for SA but the same amount of drop for LA. 

\subsubsection{Data Augmentation}
As seen in table \ref{tab:pretraining-dataaug}, a naive strategy of simply shuffling the examples from the aforementioned strategies into the first 80\% of mini batches during the fine-tuning phase did not provide significant improvements in single model performance over BERT$_{+WWM}$. This may indicate that the NQ dataset is sufficiently large so as to not require additional examples.  Instead, pre-training the base model on a similar task like SQuAD 2.0 on top of the WWM BERT model seems to be the best strategy for maximizing single model performance and outperforms the previous SOTA: a BERT model trained with $4$ million additional synthetic question answer pairs. Another interesting result is that, even the simpler (sentence shuffling) and less data intense (307,373 examples) data augmentation strategy (BERT$_{+WWM}$ w/ SOS) outperforms the previous SOTA model's use of $4$ million synthetic question answer generation model.

% \avi{please add numbers from https://arxiv.org/pdf/1906.05416.pdf 55.1F1}
% \rishav{Done}
% \avi{The table 1 is going beyond the margin..}
% \rishav{does it look OK now?} 
% \avi{still looks a tad bigger..the model names are too long.}
% \rishav{Done}

\begin{table}[]
% \small
\begin{center}
\begin{tabular}{lcc}
\toprule
 & SA F1 & LA F1 \\ \toprule
\textbf{Prior Work} \\ \toprule
DecAtt + Doc Reader & 31.4 & 54.8 \\
\cite{Parikh_2016} \\ \midrule
BERT w/ SQuAD 1.1 PT & 52.7 & 64.7 \\ 
 \cite{alberti2019bert} \\ \midrule
 BERT w/ 4M Synthetic Data & 55.1 & 65.9 \\
 Augmentation \cite{albert-synth-data} \\ \toprule
\textbf{This Work (Pre-Training)} \\ \toprule
BERT$_{WWM}$ & 55.35 & 66.04 \\ \midrule
BERT$_{SSPT}$ & 54.83 & 66.75 \\ \midrule
BERT$_{WWM}$ + SQuAD 2 PT & 56.95 & 67.28 \\ \midrule
BERT$_{WWM}$ + SQuAD 2 PT & 57.15 & 67.08 \\
+ Layer Combo\\ \midrule
BERT$_{WWM}$ + SQuAD 2 PT + AoA & \best{57.22} & \best{68.24} \\ \toprule
\textbf{This Work (Data Augmentation)} \\ \toprule
BERT$_{WWM}$ w/ SOS & 55.81 & 66.67 \\ \midrule
BERT$_{WWM}$ w/ ~21K Random & 54.05 & 66.23 \\
Examples from MRC Tasks \\ \midrule
BERT$_{WWM}$ w/ ~21K Similar & 55.18 & 66.34 \\
Examples from MRC Tasks \\ \midrule
BERT$_{WWM}$ w/ ~100K Similar & 54.68 & 65.82 \\
Examples from MRC Tasks \\ \bottomrule
\end{tabular}
\caption{Short \& long answer F1 performance of BERT-for-QA models on NQ dev. We abbreviate pre-training with PT.}
\label{tab:pretraining-dataaug}
\end{center}
\end{table}

\begin{table}[]
% \small
\centering
\begin{tabular}{lll}
\toprule
Neg Sampling Rate & Neg Sampling Rate & SA F1 \\
for Answerable &  for Un-Answerable  &  \\ \toprule
1\% & 1\% & 45.22 \\ \midrule
2\% & 2\% & 46.20 \\ \midrule
4\% & 4\% & 46.45 \\ \midrule
5\% & 5\% & 45.94 \\ \midrule
1\% & 4\% & \textbf{47.02} \\ \bottomrule
\end{tabular}
\caption{Performance on NQ dev using a preliminary \bertbase{}-for-QA model with varying sub-sampling}
\label{tab:subsampling}
\end{table}

\subsection{Ensembling}

\subsubsection{Seed Experiments}
Table \ref{tab:exp:ens:search} shows there is a benefit to ensembling multiple versions of the same model trained with different random seeds at training time. Specifically, there is a gain of roughly 2.5\% in both SA and LA F1 by ensembling four models.

\subsubsection{Multiple Model Ensembling Experiments}
\begin{table}[]
% \small
\begin{tabular}{lcc}
\toprule
  & SA F1 & LA F1  \\ \toprule
 Best Single Model & 56.14 & 67.10 \\ \toprule
 Ensemble of Best Model Trained & 58.73 & 69.61 \\
 with Random Seeds & & \\ \midrule
 Exhaustive Search (Short Answer) & 59.64 & 69.98   \\ \midrule
 Exhaustive Search (Long Answer) & 59.64 & 70.49   \\ \midrule
 Greedy (Short Answer) & 59.07 & 69.81 \\ \midrule
 Greedy (Long Answer) & \textbf{59.71} & \textbf{70.84}  \\ \bottomrule
\end{tabular}
\caption{Ensemble performance on NQ dev-test}
\label{tab:exp:ens:search}
\end{table}

As shown in table \ref{tab:exp:ens:search}, we find that ensembling a diverse set of models can provide an additional 1\% boost in SA F1 and a 1.2\% boost in LA F1 over simply ensembling the same model configuration with different random seeds during training.

Specifically, performing a greedy search and optimizing for \textit{long} answer performance appears to generalize best to the dev-test set. We hypothesize that the reasons for the superior generalization of the greedy approach over exhaustive is that exhaustive search is ``overfitting'' to the examples in dev-train.  Another potential cause of the better generalization of greedy is that it can search more candidates due to the decreased computational complexity.  

Similarly we hypothesize the reason optimizing for long answer F1 generalizes better for short and long answers is due to the strict definition of correctness for Natural Questions which requires exact span matching \cite{Kwiatkowski2019NaturalQA}.  

In our final search over all ensembles using the greedy (long answer) search, the algorithm selects an ensemble consisting of the following models: (1) BERT$_{WWM}$ + SQuAD 2 PT + AoA (2) BERT$_{WWM}$ + SQuAD 2 PT (3) BERT$_{WWM}$ + SQuAD 2 PT (4) BERT$_{SSPT}$. So only one of the chosen model configurations is that of the single best performing model.  The remaining models, though outperformed as individual models, provide a boost over multiple random seed variations of the best single model configuration.

\begin{table}[]
\centering
% \small
\begin{tabular}{lcc}
\toprule
Aggregation Strategy  & SA F1 & LA F1  \\ \toprule
Max  & \best{0.5971} & 0.7084  \\ \midrule
Reciprocal Rank Sum & 0.5728 & 0.7066  \\ \midrule
Exponential Sum  & 0.5826 & 0.7040  \\ \midrule
Noisy-Or  & 0.573 & \best{0.715} \\ \bottomrule
\end{tabular}
\caption{Performance on NQ dev-test for varying aggregation strategies for duplicate answer spans (using greedy long answer search)}
\label{tab:exp:ens:norm-agg}
\end{table}

\textbf{Duplicate Answer Span Aggregation} Table \ref{tab:exp:ens:norm-agg} shows further experimentation with the greedy long answer ensembling strategy where we vary the aggregation strategies for duplicate answer span predictions. We find that using max aggregation results in the best short answer F1 whereas using normalized noisy-or aggregation results in the best long answer F1.  Therefore, for our final submission, we use a combination strategy of producing short answer predictions using a greedy long answer search with max score for duplicate spans and long answer predictions using a greedy long answer search with noisy-or scores for duplicate spans.
\section{Conclusion}
\label{sec:conclusion}
We outline MRC algorithms that yield SOTA performance on benchmark datasets like SQuAD and show that a very simple approach involving transfer learning reaches the same performance while being computationally inexpensive. We also show that the same simple approach has strong empirical performance and yields the new SOTA on the NQ task as it outperforms a QA system trained on 4 million examples when ours was trained on only 307,373 (i.e. the size of the original NQ training set). Our future work will involve adding larger pre-trained language models like RoBERTa and XLNet.

\bibliography{main}
\bibliographystyle{aaai}

\end{document}